\documentclass[final,5p,times,twocolumn]{elsarticle}

\usepackage{manuscript/src/mypreamble}

\newacronym{ts}{$T_{s}$}{supply air temperature, $^{\circ}$C}
\newacronym{BMS}{BMS}{Building Management System}
\newacronym{HVAC}{HVAC}{Heating, Ventilation, and Air Conditioning}

\begin{document}

    \begin{frontmatter}

\title{Structural hierarchical learning for energy networks}

\author[TUeaddress,DTUaddress]{Julien Leprince \corref{mycorrespondingauthor}}
\address[TUeaddress]{Technical University of Eindhoven, 5 Groene Loper, Eindhoven 5600 MB, the Netherlands}
\cortext[mycorrespondingauthor]{Corresponding author}
\ead{j.j.leprince@tue.nl}

\author[TUeaddress]{Waqas Khan}
\ead{w.khan@tue.nl}

\author[DTUaddress]{Henrik Madsen}
\address[DTUaddress]{Technical University of Denmark, Building 303B Matematiktorvet, Lyngby 2800, Denmark}
\ead{hmad@dtu.dk}

\author[DTUaddress]{Jan Kloppenborg M{\o}ller}
\ead{jkmo@dtu.dk}

\author[TUeaddress]{Wim Zeiler}
\ead{w.zeiler@tue.nl}

\begin{abstract}
Many sectors nowadays require accurate and coherent predictions across their organization to effectively operate.
Otherwise, decision-makers would be planning using disparate views of the future, resulting in inconsistent decisions across their sectors.
To secure coherency across hierarchies, recent research has put forward hierarchical learning, a coherency-informed hierarchical regressor leveraging the power of machine learning thanks to a custom loss function founded on optimal reconciliation methods.
While promising potentials were outlined, results exhibited discordant performances in which coherency information only improved hierarchical forecasts in one setting. 
This work proposes to tackle these obstacles by investigating custom neural network designs inspired by the topological structures of hierarchies.
 Results unveil that, in a data-limited setting, structural models with fewer connections perform overall best and demonstrate the value brought by coherency information in both accuracy and coherency forecasting performances, provided individual forecasts were generated within reasonable accuracy limits.
Overall, this work expands and improves hierarchical learning methods thanks to a structurally-scaled learning mechanism extension coupled with tailored network designs, producing a resourceful, data-efficient, and information-rich learning process.
\end{abstract}


\end{frontmatter}

    

    \section{Introduction}
Optimal decision-making necessitates an accurate estimation of the future in order to effectively perform. This is true in all domains.
However, in many cases, accurate forecasts alone do not suffice. In particular, systems organized over multiple abstraction levels principally require their predictions to be consistent across all considered layers, not to result in unaligned decisions \cite{nystrup2020temporal}.
This phenomenon has become a concern in numerous domains encompassing retail \cite{kremer2016sum}, stock management \cite{spiliotis2021hierarchical}, tourism \cite{kourentzes2019cross, athanasopoulos2009hierarchical} as well as smart energy management \cite{taieb2021hierarchical, bergsteinsson2021heat}.
The latter illustrates the situation quite well while being at the center of current preoccupations toward decarbonizing existing energy systems \cite{gu2014modeling, panday2014review}.
Typically, energy networks are composed of multiple abstraction levels, divided into groups that further subdivide into groups of groups over multiple layers, referred to as hierarchies or trees.
Such structures possess varying scales of information, which are commonly employed disparately to produce element-specific forecasts across the network elements, so-called \textit{independent} or \textit{base} forecasts \cite{athanasopoulos2009hierarchical}.
Further, energy network operators now profit from high-frequency measurements at multiple levels of the system, backing accurate predictions across different aggregations, i.e., with per seconds to yearly resolutions across sub-meters to regional aggregated information \cite{taieb2021hierarchical, ahmad2020review, peng2021flexible}.
However, the diversity of these models inevitably produces inconsistencies across aggregation levels, i.e., lower-level forecasts may not add up to higher-level ones and vice-versa \cite{spiliotis2020cross}. 
The resulting challenge decision-makers are confronted with is obtaining \textit{coherent} forecasts across disparate levels of the system. A hierarchy is defined as coherent when its values at the disaggregate and aggregate scales are equal when brought to the same level \cite{kourentzes2019cross}.
Optimal decision-making consequently necessitates not only accurate but coherent forecasts in order to effectively perform.

\subsection{Securing coherency}
In this context, data-driven approaches have been proposed to produce coherent predictions from disconnected, and presumably inconsistent, base forecasts \cite{athanasopoulos2009hierarchical, hyndman2011optimal}, referred to as \textit{reconciliation} \cite{weale1988reconciliation}. The process exploits linear balancing equations from covariance compositions inherent to hierarchical time series to optimally re-adjust coherency mismatches.
Prevalent techniques encompass elemental bottom-up and top-down approaches, as well as trace minimization from optimal combinations referred to as generalized least-squares.

Bottom-up reconciliation consists in aggregating together across the hierarchy the base forecasts of the very bottom level of the hierarchy, i.e., the tree leaves, thus enforcing its coherency \cite{edwards1969should}. The approach benefits from detailed information on the most disaggregate levels of the tree, avoiding subsequent information loss from aggregation \cite{athanasopoulos2009hierarchical}. It is, however, poorly suited to time series displaying low signal-to-noise ratios as bottom-up aggregation would be unlikely to provide accurate forecasts across the upper levels of the tree \cite{hyndman2011optimal}.

Top-down reconciliation, on the other hand, proceeds to disaggregate and distribute the top-level forecast of the hierarchy, i.e., tree root, either from historical \cite{grunfeld1960aggregation} or forecasted \cite{athanasopoulos2009hierarchical} proportions of the data.
The method notably produces accurate predictions for higher aggregation levels of the tree, leveraging clear uncovered trends from aggregated time series, while being valuable for low-count data. Yet, as a result of aggregation, large quantities of information cannot be exploited, such as temporal dynamics \cite{athanasopoulos2009hierarchical}. The approach also relies on a unique top-level model, thus creating higher performance risks originating from eventual model inaccuracies or misspecification \cite{kourentzes2019another}.

Optimal reconciliation subsequently emerged as a solution to exploit the full extent of available information across a given hierarchy. It exploits correlations and covariances present in hierarchical time series, linearly reconciling base forecasts towards coherency with varied interactions between tree levels. However, covariance structures of hierarchies are challenging to estimate from base forecasts, such that even with high-frequency data available, assumptions on its form must be made \cite{doi:10.1080/01621459.2015.1058265}. 
In this context, ordinary (OLS) \cite{athanasopoulos2009hierarchical,hyndman2011optimal}, weighted (WLS) \cite{HYNDMAN201616}, and generalized least-square (GLS) \cite{wickramasuriya2019optimal} estimators were proposed providing varying approximations of the covariance matrix, namely identity, variance, and covariance respectively.
The approaches demonstrated improved results compared to other commonly adopted technics while the generalized least-square formulation guaranteed reconciled forecasts to be, in mean or in sample, as good as their original base performance.

\subsection{From hierarchical forecasting to hierarchical learning}
Hierarchical forecasting thus emerged as a solution providing coherent predictions within an established hierarchical structure.
Typically, (base) forecasts are first estimated across the tree in a disjointed manner. This allows models to be hierarchically-tailored to the considered element of the hierarchy while exploiting available node-specific information \cite{kourentzes2019cross}.
Then, forecasts are linearly combined (reconciled), securing base forecast coherency from structural hierarchical information. This phase brings formerly detached elements together by sharing targeted information between aggregation levels to generate coherent predictions.

Undertaking these two phases separately, however, engenders two noteworthy shortcomings; it inherently deprives similar forecasting models from 
(\textit{i}) the benefits of data information (learning) transfer across models,  as well as
(\textit{ii}) structural hierarchical information together with coherency requirements which cannot be exploited by forecasting models despite their valuable information.

Thereby, recent work suggested bridging forecasting with the coherency requirements of the reconciliation phase \cite{https://doi.org/10.48550/arxiv.2301.12967}. 
The approach proposes (\textit{i}) a unique hierarchical forecasting model, providing a global overview of information across the hierarchy to the regressor, while (\textit{ii}) including coherency requirements within its learning process. Notably, it puts forward a custom loss function, embedding optimal reconciliation information in the learning process of deep neural networks founded on established field taxonomy.
Overall, the work expanded and united traditionally disjointed methods together unveiling a novel generation of forecasting regressors: \textit{hierarchical learners}.

While the work portrays a pioneering and promising method for hierarchical learning, results, however, displayed disparate performances, in which coherency learning was found to improve forecasting scores in simply one setting. 
The analysis particularly unveiled two important challenges the approach struggled with; laborious learning and faulty coherent learning from scaled trees.
Indeed, while unifying hierarchical predictions under one machine-learning model fuels the regressor with a rich data-learning process, the consequent size of the produced model renders the learning process of the algorithm arduous. The produced number of weights to update possesses a $n\cdot m$ relationship between layers of size $n$ and $m$, which exponentially explodes as the hierarchy grows larger.
Additionally, estimated forecasts of large hierarchical multi-output regressors might produce either conflicting or highly correlated outputs, that are challenging to capture from limited data.
Further, hierarchical-coherent learning exhibited faulty performances in certain settings, where top-level forecasts mirrored their expected values in the negative domain. The identified undesirable result was traced back to the normalization of hierarchical time-series required for adequate learning, which was shown to irregularly bias the coherency learning process.

In light of this, tailored designs need to be examined to exploit the full potential of hierarchical learning methods.

\subsection{Motivation}
This paper intends to address identified complications by investigating custom neural network designs mirroring the structural dimensions of hierarchies. 
Namely, we propose first to divide large hierarchical models into separate partitions, thus effectively reducing the number of weights to learn for a more data-efficient training process while producing distinct model components tailored to node-specific elements. 
Second, we connect separated model partitions leveraging established structural hierarchical designs, hence incorporating structural information accurately into the model layout while allowing separated models to share specific information with one another. 
Finally, leveraging batch normalization within the design of neural networks, we introduce a learning process robust to the scale differences endowed with hierarchies, avoiding a priori tree normalization and its identified subsequent biased coherency learning.
Contributions of this work can ergo be summarized as three-fold;
\begin{enumerate}
    \item We propose to expand promising hierarchical learning regressors by putting forward a novel \textit{structural} hierarchical learning method. Founded on tailored network designs, the approach exploits topological hierarchical information from trees targeted to support a resourceful, data-efficient, and information-rich learning process.
    \item We formally evaluate the relative performance brought by the addition of the coherency requirement across all examined designs, thus clearly establishing the realized value of coherent hierarchical learning.
    \item Finally, we open-source all developed implementations backing effective knowledge dissemination and allowing full research reproducibility under a public GitHub repository \footnote{\href{https://github.com/JulienLeprince/structuralhierarchicallearning}{https://github.com/JulienLeprince/structuralhierarchicallearning}}. 
\end{enumerate}

With these contributions, this work adequately addresses the exposed challenges of hierarchical learning. 
By putting forward tailored, ingenious architectures of neural networks it effectively reduces hierarchical model complexities while serving improved and coherency-aware hierarchical forecasts.
The work notably serves domains such as stock management, retail, business analytics, and distribution networks, by providing improved and more consistent forecasts across all levels of considered hierarchies.

The rest of this paper is organized as follows; the mathematical foundations for the reconciliation of hierarchical time-series is first presented along with hierarchical learning regressors and their structural extension in Sec. \ref{sec:hl}. Next, the implementation specifics related to the machine learning design and evaluated case study are exposed under Sec. \ref{sect_implementation}, followed by its results in Sec. \ref{sect_res}. Section \ref{sect_con} concludes the article.
    
    \section{Structural hierarchical learning}\label{sec:hl}
Coherency requirements and reconciliation technics feeding hierarchical learning regressors are first presented, serving as a foundation on which topologically-tailored model designs can be built.

\subsection{Hierarchies and coherency}
\begin{figure}
    \centering
    \begin{adjustbox}{width=0.74\linewidth}
        \includegraphics{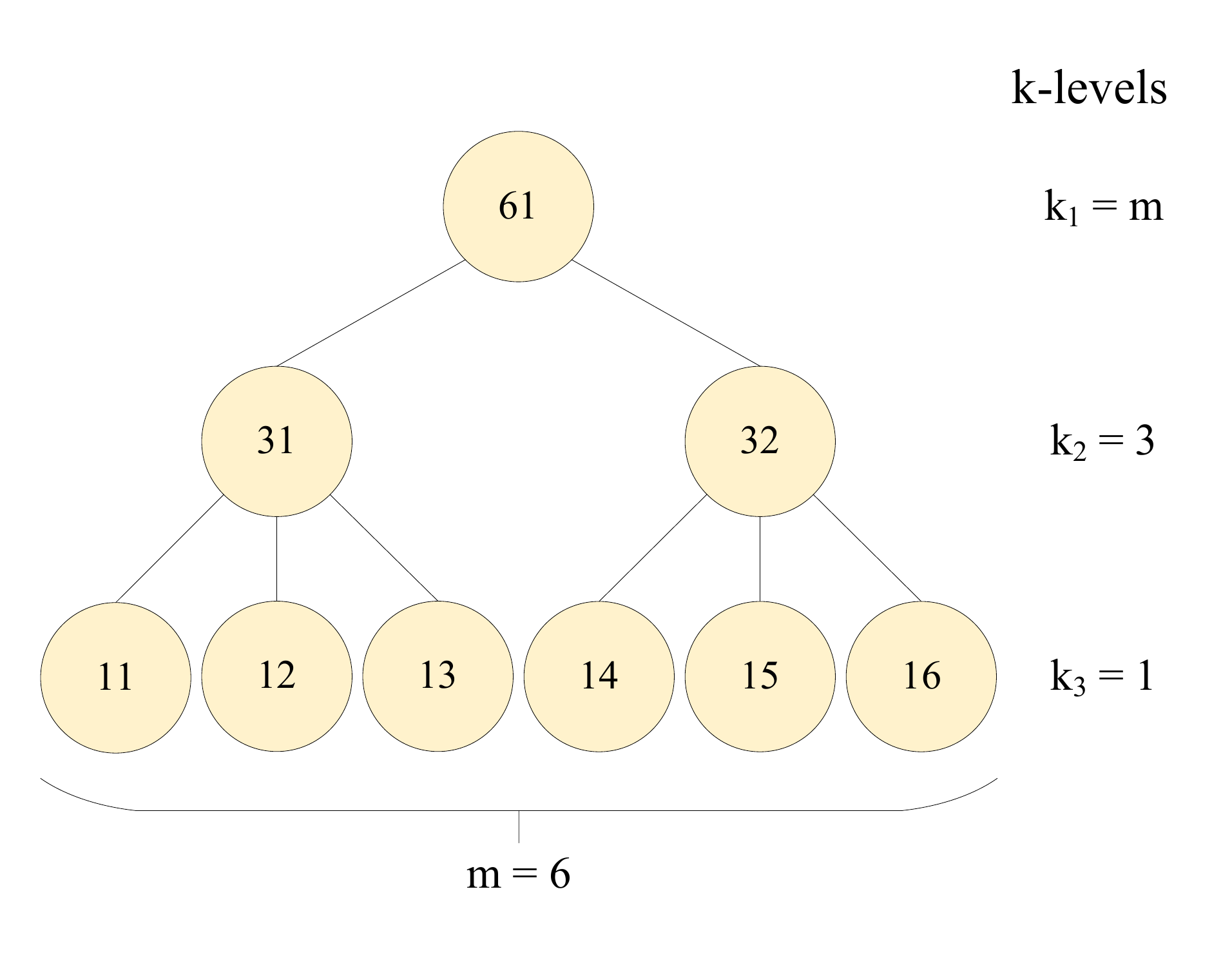}
    \end{adjustbox}
    \caption{A two-level hierarchical tree diagram.}
    \label{fig:tree}
\end{figure}

Let us introduce the simple hierarchy of Fig. \ref{fig:tree} to illustrate the requirements for coherency.
Every element (node) of the hierarchy can be labeled as $y_{kj}$, where subscripts \textit{k} and \textit{j} stand for the aggregation-level and node observations respectively.
We define $k_1=m$ as the most aggregate level of the hierarchy (tree root), i.e., node $y_{61}$, and $k_3=1$ as the most disaggregate level (tree leaves), i.e., nodes $y_{1j}$ where $j \in [1:m]$.
In such a setting, two important components must be considered; the number of leaves, denoted as \textit{m}, and the total number of nodes on the tree \textit{n}. Here $n=9$ and $m=6$.

Stacking all hierarchical elements in a \textit{n}-dimensional vector $\boldsymbol{y}=(y_{61}, y_{31}, y_{32}, y_{11}, y_{12}, y_{13}, y_{14}, y_{15}, y_{16})^T$, and bottom-level observations in an \textit{m}-dimensional vector $\boldsymbol{b}=(y_{11}, y_{12}, y_{13}, y_{14}, y_{15}, y_{16})^T$, we can write
\begin{flalign}
    \boldsymbol{y} = S \boldsymbol{b} \text{,}
\end{flalign}
where $S$ is the summation matrix, here expressed as
\begin{flalign}
    S = 
    \begin{bmatrix}
    1 & 1 & 1 & 1 & 1 & 1 \\
    1 & 1 & 1 & 0 & 0 & 0\\
    0 & 0 & 0 & 1 & 1 & 1 \\
    \multicolumn{6}{c}{I_m} \\
    \end{bmatrix}
     \text{,}
\end{flalign}
which is of dimension $n \times m$, and $I_m$ is an identity matrix of size $m$.
$S$ encapsulates the hierarchical structure of the tree, from which the complete hierarchy $\boldsymbol{y}$ can be obtained given the tree leaves $\boldsymbol{b}$. 
Consider how $S$ captures the coherency requirements within the hierarchy, included here as the linear summations of the bottom-level observations.
By introducing a matrix
\begin{flalign}\label{eq:G}
    G = \big[ 0_{m \hspace{.5mm}\times\hspace{.5mm} (n-m)} \hspace{.5mm}|\hspace{.5mm} I_m \big] \text{,}
\end{flalign}
of order $m \hspace{.5mm}\times\hspace{.5mm} n$ that extracts the $m$ bottom-level forecasts, the reconciliation constraint is formulated as
\begin{flalign}\label{eq:SG}
    \tilde{\boldsymbol{y}} = SG\tilde{\boldsymbol{y}} \text{.}
\end{flalign}
Reconciliation is necessary when base forecasts $\hat{\boldsymbol{y}}$ do not satisfy this constraint \cite{nystrup2020temporal}. In such situations, Eq. \eqref{eq:SG} becomes $\tilde{\boldsymbol{y}} = SG\hat{\boldsymbol{y}}$,
where $G$ maps the base forecasts into the reconciled tree-leaves and $S$ sums these up to a set of coherent forecasts $\tilde{\boldsymbol{y}}$. $SG$ can thus be thought of as a reconciliation matrix taking the incoherent base forecasts as input and reconciling them to $\tilde{\boldsymbol{y}}$.
This formulation, however, is sub-optimal as $G$, as defined in Eq. \eqref{eq:G}, only considers information from a single level.

To include the exploitation of all aggregation levels in an optimal manner, Hyndman et al. \cite{hyndman2011optimal} and later, Van Erven and Cugliari \cite{10.1007/978-3-319-18732-7_15} and Athanasopoulos et al. \cite{ATHANASOPOULOS201760} formulated the reconciliation problem, as linear regression models. Exploiting a defined hierarchical structure, reconciled forecasts are found employing the generalized least-squares problem:
\begin{equation}\label{eq:optimalrecon2}
\begin{aligned}
    \text{minimize} &\hspace{5mm} \big(\tilde{\boldsymbol{y}} - \hat{\boldsymbol{y}}\big)^T \Sigma^{-1} \big(\tilde{\boldsymbol{y}} - \hat{\boldsymbol{y}}\big) \text{,}\\
    \text{subject to} &\hspace{5mm} \tilde{\boldsymbol{y}} = SG\tilde{\boldsymbol{y}} \text{,}
\end{aligned}
\end{equation}
where $\tilde{\boldsymbol{y}} \in \mathbb{R}^n$ is the decision variable of the optimization problem and $S \in \mathbb{R}^{n \times m}$ and $G  \in \mathbb{R}^{m \times n}$ are constant matrices defined by the structure of the hierarchy. The parameter $\Sigma \in \mathbb{R}^{n \times n}$ is the positive definite covariance matrix of the coherency errors $\varepsilon = \tilde{\boldsymbol{y}} - \hat{\boldsymbol{y}}$, which are assumed to be multivariate Gaussian and unbiased, i.e., with zero mean.

If $\Sigma$ were known, the solution to \eqref{eq:optimalrecon2} would be given by the generalized least-squares (GLS) estimator
\begin{flalign}\label{eq:sigma}
    \tilde{\boldsymbol{y}} = S \big(S^T \Sigma^{-1} S \big)^{-1} S^T\Sigma^{-1}\hat{\boldsymbol{y}} \text{,}
\end{flalign}
which has been employed in close to all notable hierarchical forecasting works over the last years \cite{nystrup2020temporal, kourentzes2019cross, athanasopoulos2009hierarchical, taieb2021hierarchical, spiliotis2020cross, hyndman2011optimal, wickramasuriya2019optimal, ATHANASOPOULOS201760}.
The precision matrix $\Sigma^{-1}$ is used to scale discrepancies from the base forecasts and can also be referred to as the weight matrix.
The recurrent challenge in estimating $\Sigma^{-1}$ originates from its dimension $n \times n$ which can potentially become very large.

Varying approximations of the covariance matrix were consequently put forward in previous work, including identity (ordinary least-square), structural \cite{ATHANASOPOULOS201760}, variance \cite{nystrup2020temporal,ATHANASOPOULOS201760}, Markov \cite{nystrup2020temporal}, and spectral \cite{nystrup2021dimensionality} scaling, along with shrank covariance estimations \cite{nystrup2020temporal,wickramasuriya2019optimal}. 
All of these allow distinct information sharing between elements of the hierarchy serving optimal reconciliation in differing settings.

\subsection{Hierarchical learning}
The more recently proposed hierarchical learning method \cite{https://doi.org/10.48550/arxiv.2301.12967} intended to unify hierarchical forecasts together while exploiting the coherency requirement of produced predictions. 
The approach builds on the presented formulations of optimal reconciliation, see Eq. \eqref{eq:sigma}, included in machine learning processes thanks to a custom loss function.

To propose a regressor that is robust to scale differences and avoids normalization requirements and its subsequent identified faulty-coherency learning \cite{https://doi.org/10.48550/arxiv.2301.12967}, we leverage the  Mean Structurally-Scaled Square Error (MS3E) defined in Ref. \cite{https://doi.org/10.48550/arxiv.2301.12967} for the definition of the learning loss function.
Structurally scaled errors must, indeed, be considered in hierarchical forecasting as produced predictions will possess major scale differences inherent to hierarchical structures.
Designing hierarchical loss functions notably differentiates itself in this way from the scaled multi-output regressors of Ref. \cite{https://doi.org/10.48550/arxiv.2301.12967}. To avoid a biased learning process favoring the top levels of the aggregation, where predicted values possess larger magnitudes, such structural scale differences must be adjusted.
To consider this, we employ structural scaling leveraging the aggregation-level vector 
\begin{flalign}\label{eq:kappa}
    \boldsymbol{\kappa}_{str} = S\boldsymbol{1}_m \text{.}
\end{flalign}
For example, for the hierarchy illustrated in Fig. \ref{fig:tree}, $\boldsymbol{\kappa}_{str}$ = (6, 3, 3, 1, 1, 1, 1, 1, 1).
The structural-hierarchical loss function $\mathcal{L}^{h}$ can then be expressed as
\begin{flalign}\label{eq:Lsh}
    \mathcal{L}^{sh}\big(\mathcal{Y}, \widehat{\mathcal{Y}}|\Theta\big) = \frac{1}{T}\sum^{T}_{t=1} \Big( (\boldsymbol{y}_t - \hat{\boldsymbol{y}}_t) \oslash \boldsymbol{\kappa}_{str} \Big)^2  \text{.}
\end{flalign}
where $T$ is the number of time-steps, $\mathcal{L}^{sh}$ denotes the mean structurally-scaled square loss function between the predicted independent \textit{base} forecast set $\widehat{\mathcal{Y}}$ subject to a set of parameters $\Theta$ and a set of observed values $\mathcal{Y}$. The operator $\oslash$ is a Hadamard division.
It should be noted that while the proposed hierarchical forecasting loss function employs the MS3E metric, any other accuracy metric can be exploited provided they use structurally scaled error $\boldsymbol{e}^{str}_{t} = (\boldsymbol{y}_t - \hat{\boldsymbol{y}}_t) \oslash \boldsymbol{\kappa}_{str}$ as error reference.


The structural-coherency loss function is formulated as the structurally scaled differences between predicted values $\hat{\boldsymbol{y}}$ and their reconciled counterpart $\tilde{\boldsymbol{y}}$, following the reconciliation product of Eq. \eqref{eq:sigma}.
The coherency error $\boldsymbol{e}^{coh}_{t}$ and subsequent scaled loss function $\mathcal{L}^{sc}$ can consequently be expressed as
\begin{flalign}\label{eq:Lsc}
    \boldsymbol{e}^{coh}_{t} = \hat{\boldsymbol{y}}_t - S \big(S^T \Sigma^{-1} S \big)^{-1} S^T\Sigma^{-1}\hat{\boldsymbol{y}}_t \text{,} \\
    \mathcal{L}^{sc}\big(\mathcal{Y}, \widehat{\mathcal{Y}}|\Theta\big) = \frac{1}{T}\sum^{T}_{t=1} \Big( \boldsymbol{e}^{coh}_{t} \oslash \boldsymbol{\kappa}_{str} \Big)^2 \text{.} \label{eq:cohrency_loss}
\end{flalign}

Both structural accuracy and coherency losses are then combined together forming the structural-hierarchical-coherent loss function $\mathcal{L}^{shc}$,
\begin{flalign}\label{eq:Lshc}
    \mathcal{L}^{shc}\big(\mathcal{Y}, \widehat{\mathcal{Y}}|\Theta\big) = \alpha \mathcal{L}^{sh}_t + (1-\alpha) \mathcal{L}^{sc}_t  \text{,}
\end{flalign}
where $\alpha \in [0,1]$ weights the structural hierarchical loss against the structural coherency loss. This avoids the over-adjustment of weights during the training of the regressor due to the addition of the coherency loss to the loss function. We typically set $\alpha$ to 0.75 for structural hierarchical learning.

\subsection{Structural learning}
Hierarchical learning regressors are here tailored to exploit topological structures of hierarchies, allowing the design of bespoke \textit{structural} hierarchical models.
This is accomplished by manipulating two distinct characteristics of machine learning models, namely, neuron partitions and network weights connecting partitions together.
This work proposes to ingeniously shape both features echoing common hierarchical attributes.

\subsubsection{Partitioning hierarchical models}
\begin{figure}
    \centering
    \begin{adjustbox}{width=0.98\linewidth}
        \includegraphics{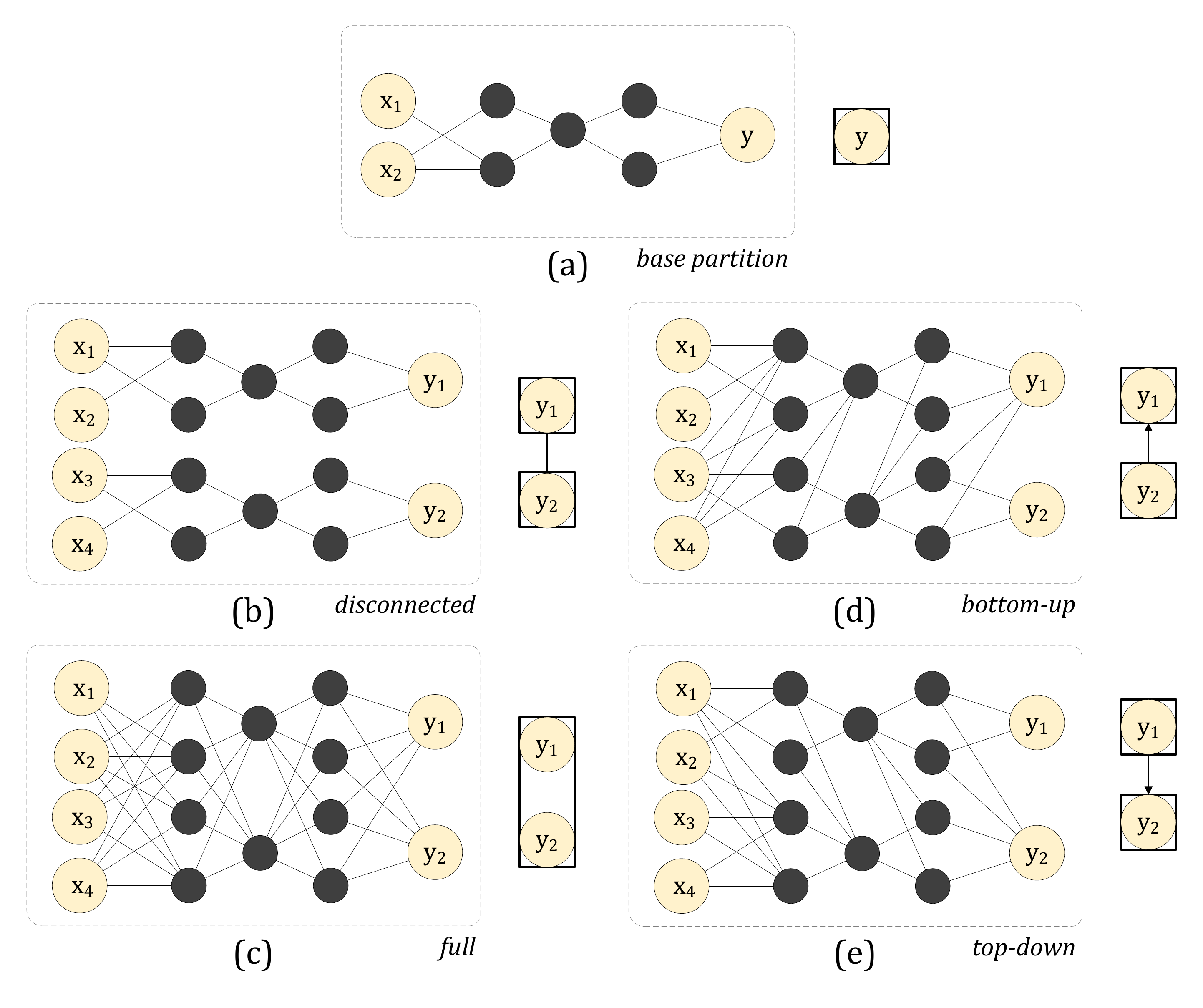}
    \end{adjustbox}
    \caption{Example illustrations of (a) \textit{base} single-task fully-connected deep neural network, composed 3 hidden layers and 2 inputs, followed by four multi-task stacked variations of the former model where $y_1$ and $y_2$ relate to parent and child nodes of a larger hierarchical ensemble respectively.
    The multi-task models encompass (b) \textit{disconnected}, (c) \textit{full}y-connected, (d) \textit{bottom-up}, and (e) \textit{top-down} partition links, creating a deep neural network, composed of 3 similar hidden layers, 4 inputs, and 2 outputs.}
    \label{fig:abcbespoke}
\end{figure}

We begin by defining a single-task fully connected deep neural network of 3 hidden layers and 2 inputs, as illustrated by the (a) \textit{base} model of Fig. \ref{fig:abcbespoke}.
To best conceptually illustrate network partitions, we propose to compare two stacked variations of said single-task regressor (a), thus forming the multi-task regressors (b) and (c). While both models possess similar number of layers, neurons, inputs, and outputs, the weights connecting these elements into a common model are here notably different. In model (c) all neurons composing the hidden layers are connected to their preceding and succeeding elements, i.e., input, hidden layer, or output, while in model (b) two distinct \textit{partitions} can be recognized formed by their parent regressor (a).
Naturally, in a common prediction task, it would not be considered valuable to assemble two disconnected models together to form model (b). There would be no performance or computational gains to expect from such a setup compared to employing two independent models (a) instead for instance. 
In a hierarchical forecasting setting, however, it becomes on the contrary rather beneficial to gather produced outputs under the hood of one model. Indeed, in this way coherency requirements of produced predictions can be exploited by leveraging the coherency loss function.
Additionally, it becomes clear that producing specific partitions over a defined neural network considerably reduces the number of weights to update. 
Taking, again, the illustrated example of Fig. \ref{fig:abcbespoke}, model (c) possesses a total of 36 weights to update, while model (b) has only 20. 
This significant reduction in the number of weights to learn can serve two desirable outcomes.
First, fewer weights to learn implies fewer iterations required to calibrate the model and subsequently fewer data instances, an entity that is typically expensive to gather in large and qualitative quantities.
And, second, it can effectively support targeted single-task learning by helping isolate eventual conflicting multi-output predictions from one another in the model design.

We propose three partition variations, serving as alternatives to the fully-connected hierarchical model of Ref. \cite{https://doi.org/10.48550/arxiv.2301.12967}, echoing established tree topologies, namely, fully-disconnected, leaf-connected, and k-level partitions, which we visually introduce in Fig. \ref{fig:all_NN_designs} as vertical lattices.
Fully disconnected partitioning intends on mirroring independent, or base, forecasting by considering each tree node as a disjointed partition of the neural network. Similarly to the (b) model of Fig. \ref{fig:abcbespoke}, tree-node partitions are stacked together to form a larger, hierarchical model. We refer to this partitioning as \textit{tree}.
Leaf-connected partitioning, referred to as \textit{cutree}, suggests linking leaf elements with identical parents together while considering the rest of the tree nodes as independent partitions. This setup is particularly interesting for hierarchies built from time-series clustering, where leaf elements typically display similar dynamics.
Finally, k-level partitioning proposes to group hierarchical elements possessing alike aggregation levels into common partitions. The advantage of this approach is granting time series with similar levels of aggregation, and subsequent signal-to-noise ratios, a shared partition to capture these, possibly similar, dynamics. We refer to this partitioning as \textit{klvl}.

\subsubsection{Creating topological bridges}\label{subsubsect:topology}
While leveraging distinct node-specific models in hierarchical forecasting is profitable, it becomes relevant to create bridges between separated layers, allowing targeted and effective learning across defined hierarchical modeling partitions.
Two topological bridges have already been presented, namely disconnected partitioning, or \textit{disc}, and fully connected ones.

To complement these we propose two well-established hierarchical connections, namely, (d) bottom-up and (e) top-down illustrated in Fig. \ref{fig:abcbespoke}, to bridge the disconnected partitions of $y_1$ and $y_2$ together. We here assume $y_1$ and $y_2$ to be parent and child nodes within a larger hierarchical ensemble respectively.
The bottom-up setup connects the lower levels of the hierarchy with their closest higher-level elements. This implies creating neuron links, or weights, between each element of lower-level partitions, $y_2$, and their next respective parent partition, $y_1$. We refer to this partitioning as \textit{bu}.
Top-down topological bridging works in the opposite way. It creates connections between higher-level elements of the hierarchy and their associated children, i.e., here connecting $y_1$ partition elements to its child $y_2$. This partitioning is thereafter mentioned as \textit{td}.
Lastly, a subsequent topological bridge combining the above methods is put forward, i.e., bottom-up and top-down, or \textit{butd}.

\begin{figure}
    \centering
    \begin{adjustbox}{width=0.99\linewidth}
        \includegraphics{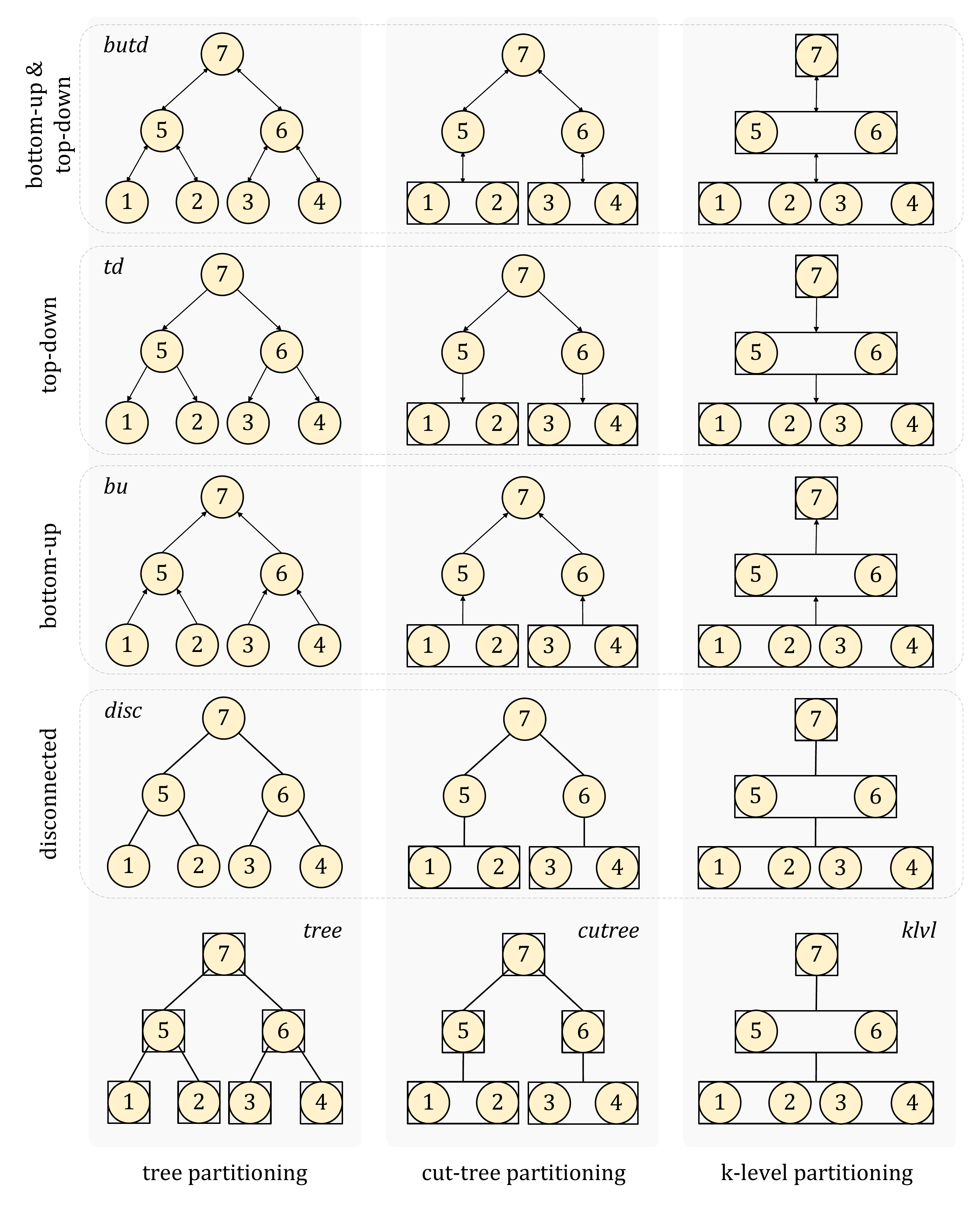}
    \end{adjustbox}
    \caption{Schematic illustration of introduced model partitions (vertical) and topological bridges (horizontal) on a two-level hierarchy}
    \label{fig:all_NN_designs}
\end{figure}

Figure \ref{fig:all_NN_designs} visually summarizes all of the introduced partitions and topological bridges, exemplified on a two-level hierarchy.


    \section{Implementation}\label{sect_implementation}
This section details the implementation-related details of our study, namely, the considered hierarchical time-series and predictive-learning setup.

\subsection{Hierarchical time-series}\label{sect_casestudies}
Our study considers a open large data set of building smart-meter measurements, namely the Building Data Genome project 2 (BDG2) \cite{miller2020building}.
This open set was selected to allow reproducibility of our method while putting forward an initial benchmark for hierarchical forecasting in the building sector. The BDG2 includes 3053 energy meters from 1636 non-residential buildings grouped by site located in Europe and, principally, North America. The set covers two full years (2016–2017) at an hourly resolution with multi-meter building measurements paired with site weather data.
Due to the heavy computational workloads arising from hierarchical forecasts, we select the \textit{Fox} site of the BDG2 for the remainder of this study. The site is composed of 133 meter-readings, which is deemed large enough to appreciate the aggregation of large numbers of building electrical loads, while not developing into too computational burdensome forecasting models.

\begin{figure}
    \centering
    \begin{adjustbox}{width=0.99\linewidth}
        \includegraphics{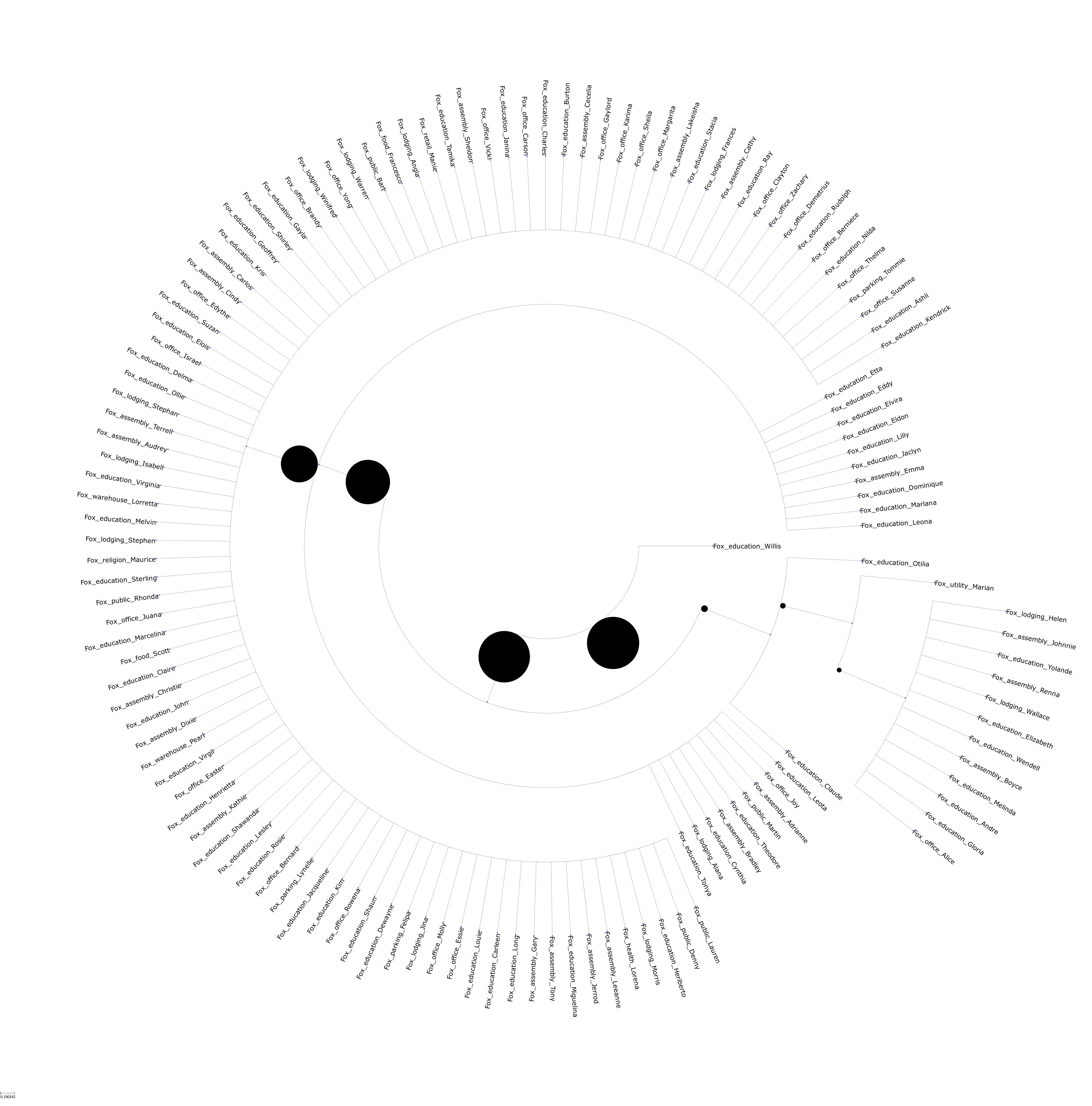}
    \end{adjustbox}
    \caption{Spatial tree structure of the \textit{Fox} site of the BDG2 dataset}
    \label{fig:Fox_tree}
\end{figure}

A spatial hierarchical structure is defined from the hierarchical clustering of the prediction target time series, i.e., electricity demand.
Buildings are consequently clustered employing the Ward variance minimization algorithm \cite{mullner2011modern}, which iteratively groups similar buildings together by minimizing the variance of the euclidean distances between their electrical loads, resulting in a large hierarchical structure of 265 nodes.
The obtained hierarchy is reduced in size by \textit{cutting the tree} using a defined distance threshold over visual inspection of the derived dendrogram. In this way, hierarchical structures located below the threshold will be clustered together, thus, effectively reducing the number of nodes to 140. 
Figure \ref{fig:Fox_tree} illustrates the attained reduced tree of the \textit{Fox} site, resulting in a power distribution-like structure, where multiple buildings are typically connected to similar buses over the low-voltage grid.

\subsection{Model learning setup}
We proceed to resample the time-series to hourly intervals. 
Time-series with cumulative missing values larger than 2 hours are removed, and smaller gaps are interpolated via a moving average using a window size of 8 hours.
We set the prediction horizon to one time-step (hour) ahead.

\subsubsection{Feature engineering}
Features are selected based on their Maximum Information Coefficient (MIC) \cite{reshef2011detecting} computed in relation to the electrical load learning target. MIC is a powerful indicator that captured a wide range of associations both functional and not while providing a score that roughly equals the coefficient of determination (R\textsuperscript{2}) of the data relative to the regression function. It ranges between values of 0 and 1, where 0 implies statistical independence and 1 a completely noiseless relationship. The advantage of using MIC for feature engineering over the more commonly employed pearson correlation indicator \cite{miller2019s} is that it captures non-linear relationships present in the data, which deep-learning models are popularly capable of detecting.  We retain features exhibiting MIC values higher than 0.25, as electric loads can typically become quite volatile and impede MIC values with noise.

Additionally, to feed the learner with the most relevant historical information of the predicted target, we select the 3 top auto-correlation values per temporal aggregation level above 0.25 as model input features. If no target auto-correlation value is above 0.25, we consider the most recent historical information, i.e., $t$ where $t+1$ is the predicted time-step and the forecast horizon is 1 hour.

Both MIC and autocorrelation selection thresholds are settings that should typically be included in the hyper-parameter optimization of the model validation phase. 
This work considers the tuning of these thresholds, however, to lay outside its scope as these rapidly become excessively burdensome.

\subsubsection{Data partitioning and covariance matrix estimation}\label{subsect:data_partitioning}
Training and testing sets are constructed employing a times-series cross-validator of the {\tt sklearn} package \cite{scikit-learn}, i.e., TimeSeriesSplit, with equal test-size in a rolling window setup.
The data set is divided into 10 batches, over which the model is trained and tested from according sets. 
We employ the heterogeneous variance approximation of the covariance matrix, which displayed good prediction performances in Ref. \cite{https://doi.org/10.48550/arxiv.2301.12967}. The heterogeneous variance includes separate variance estimates for each node. With the example hierarchy of Fig. \ref{fig:tree} this gives $\Sigma_{hvar} = \text{diag}({\sigma}^2_{61}, {\sigma}^2_{31}, {\sigma}^2_{32}, {\sigma}^2_{11}, \ldots , {\sigma}^2_{17})$.
The covariance matrix is recursively estimated in the test sets.
For the first batch training, we employ the identity covariance estimate $id$ as no forecasts are yet available. Each batch training $i$ then comes with a new covariance matrix estimate $\Sigma_i$ that is employed in the coherency loss function of the next training set $i+1$. This setup echoes the adaptive covariance matrix estimation proposed by \cite{bergsteinsson2021heat} employed for temporal hierarchies, anchored here quite organically in the learning process of neural networks.

\subsubsection{Regressor design}\label{subsubsec:modeldesign}
To build onto the design of Ref. \cite{https://doi.org/10.48550/arxiv.2301.12967}, we pursue the implementation of structural hierarchical regressors with deep neural networks, here designed according to three predominant features: partition widths, sequential layer depth, and topological bridges (weights).

Each partition is designed as a series of sequential layers decreasing proportionally in size, from the defined input layer width $w_{1p}$ to the desired output dimension $w_D$, such that
\begin{flalign}
    w_{ip} = (w_{1p} - w_{Dp}) \cdot \frac{i}{D} \text{ ,}
\end{flalign}
defines a partition's width in function of its design depth $D$. 
The subscripts $i$ and $p$ stand for the sequential layer depth and sequential layer index respectively where $i \in [1, ..., D]$ and $p \in [1, ..., P]$.
Aggregating the partitions together in the regression model then produces $n$ forecasts, ensuring that $\sum_p w_{Dp} = n$.
We select the aggregated number of features per partition as the input layer width  $w_{1p}$.

Between each sequential partition, we further introduce batch normalization and dropouts, both serving different purposes.
Batch normalization is a technique to standardize activations in intermediate layers of deep neural networks across mini-batches. It has demonstrated improved accuracies and faster convergences due to its stabilization of the learning process \cite{bjorck2018understanding}.
Additionally, introducing batch normalization allows the in and outputs of the regression model to remain unscaled, thus retaining the hierarchical structure of the coherency-loss function. 
This is an essential design improvement from Ref. \cite{https://doi.org/10.48550/arxiv.2301.12967}, which allows the tackling of observed faulty-coherency learning engendered from scaled trees.
Dropout is a technique introduced by N. Srivastava et al. \cite{srivastava2014dropout} designed to prevent overfitting by combining exponential numbers of combinations of neural network architectures efficiently. The term “dropout” refers to dropping out units of a neural network. 
Dropped-out units are removed from the network, along with all their incoming and outgoing connections, thus producing a thinned network.
In essence, dropout simulates model assembling without creating multiple networks \cite{garbin2020dropout} while increasing convergence time.

Topological bridges are then established between neurons of initially disconnected partitions following the presented connections of Sec. \ref{subsubsect:topology}, namely disconnected (\textit{disc}), bottom-up (\textit{bu}), top-down (\textit{td}), and bottom-up top-down (\textit{butd}).

The optimal number of layers of the model is selected heuristically based on prediction performances while increasing step-wise the network's depths starting from shallow 1-layer perceptrons. 
This allows the selected architecture to serve an "as simple as possible yet as complex as necessary" design. 
We select similar hyper-parameters as Ref. \cite{https://doi.org/10.48550/arxiv.2301.12967}, resulting in 3 layers, leveraging sigmoid activation functions and dropout ratios of 0.2 on all but the last layer favoring a linear activation and no dropouts. The retained $\alpha$ coefficient value is 0.75.
The presented models of Sec. \ref{sec:hl} were implemented in Python using the {\tt TensorFlow} package \cite{tensorflow2015-whitepaper}.

    \section{Results and discussion}\label{sect_res}
We describe and discuss the outcome of the implementation here. 
In particular, we evaluate the accuracy and coherency of the forecasted building loads outlined in varying heatmaps allowing insights into the performances of the forecast across the tree and forecasting methods.
The improvement ratio brought by the coherency loss function is also highlighted both for accuracy and coherency forecast performances.

\subsection{Forecast accuracy}
\begin{figure*}
    \centering
    \begin{adjustbox}{width=0.99\linewidth}
        \includegraphics{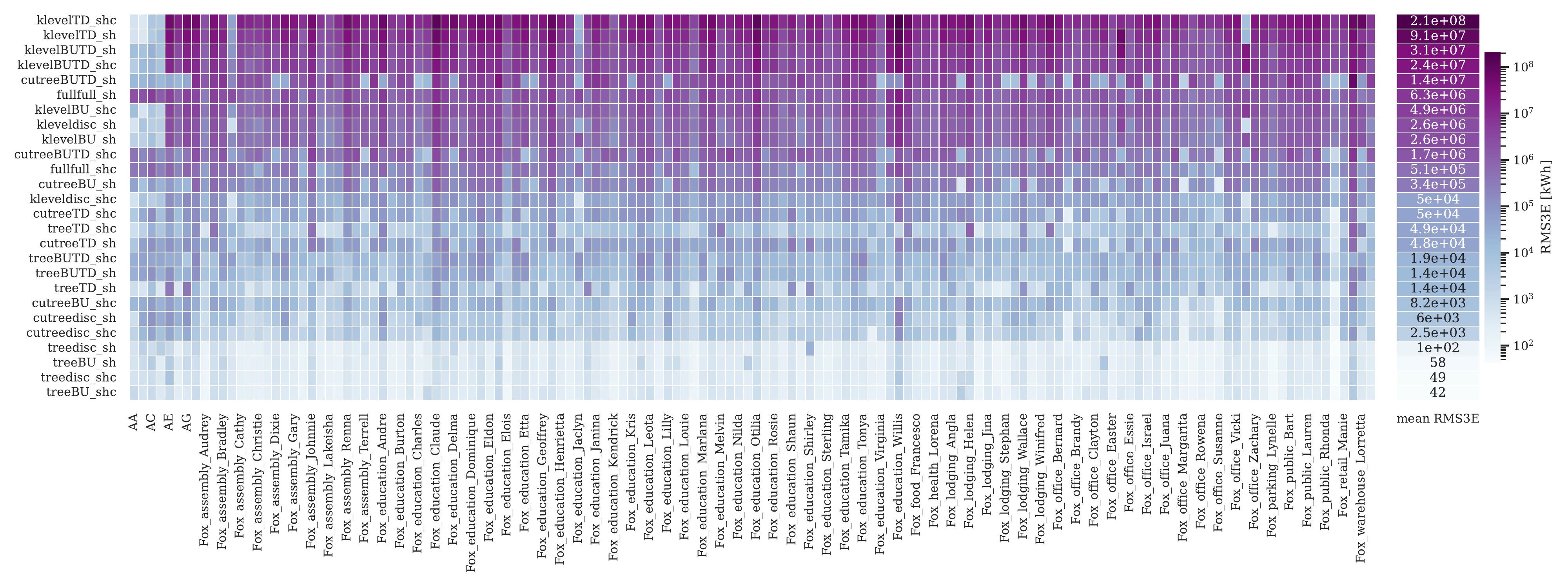}
    \end{adjustbox}
    \caption{Heatmap of the accuracy performance in Root Mean Structurally-Scaled Square Error (RMS3E) across tree nodes and forecasting methods. Structural hierarchical forecasting methods described against the y-axis are designated by their respective partition, topological bridge, and considered loss function, and are here sorted according to their forecasted accuracy.}
    \label{fig:heatmap_acc}
\end{figure*}

Figure \ref{fig:heatmap_acc} presents the forecasted accuracy of all evaluated methods over the tree nodes, sorted by their performances across the overall hierarchy.
Both extreme values of the heatmap present the tree partitioning with bottom-up (\textit{bu}) connections and structural hierarchical-coherent loss function (\textit{shc}) as  the better performer across the forecasting methods, while the k-level partitions with top-down (\textit{td}) topological bridges and \textit{shc} loss function performs the worst, by an impressive 8 order of magnitude RMS3E difference.
The notable better performers possess RMS3Es ranging from 42 to 100 kWh and all bear tree partitions that are either bridged in a disconnected (\textit{disc}) or \textit{bu} fashion. These two leading contenders each perform best with the inclusion of the coherency requirement in the loss function, i.e., \textit{shc} versus \textit{sh}.
On the other end of the heatmap, we can regroup flawed performers ranging from 3.4e5 to 2.1e8 kWh RMS3E. The structural characteristics of these networks display k-level, cut-tree, and full partitions coupled to varying topological bridges, mostly \textit{td} and bottom-up-top-down (\textit{butd}).

The tendencies that can be extracted from Figure \ref{fig:heatmap_acc} expose that (\textit{i}) structural models with fewer connections perform overall better than models with larger numbers of connections, and (\textit{ii}) that within good performers, the inclusion of coherency information in the loss function improves the performance of the overall accuracy of the forecast.

Indeed, considering the number of connections per topological-design places the tree partition as the one with the least amount of connections, followed by cut-tree, k-level, and full partitions. Inter-partition connections follow the logical increasing ordering of disconnected, \textit{bu}/\textit{td}, and \textit{butd}.
It is consequently observed that tree partitions perform best as they result in narrower layers compared to cut-tree and k-level ones. However, the least connected model design, \textit{tree-disc}, stands as the second best performer, thus demonstrating that some amount of information exchange between hierarchical layers, here \textit{bu}, is valuable for the performance of the forecast.
The flawed performers exhibit similar inclinations, where k-level partitions perform overall worse than cut-tree ones, which disregard the connections between leaves of dissimilar parents, thus cutting down their numbers.
Then, the inclusion of coherency information in the learning mechanism of the regressor produces improved forecasts for the better half of the models, with the exception of a few cases, namely \textit{td} and \textit{butd} trees. This will be further discussed under Sec. \ref{subsect:coherencyval}.

It can be noticed that the \textit{td} connections systematically perform much worse, by at least a RMS3E order of magnitude, than their \textit{bu} counterparts, i.e., within similar layer partitionings and loss functions. 
The only exception that ignores this observation is the cut-tree partitioning with \textit{sh} loss.  
This poorer performance of the  \textit{td} connection also seems to negatively impact the performance of its derivative \textit{butd}. 
In turn, the \textit{butd} linkage exclusively performs worse than its \textit{bu} setup, in similar neural network designs.
This topological bridge design, indeed, suffers from the influence of meager \textit{td} performances coupled with greater numbers of weights to learn, in a data-limited setting.

Lastly, a few peculiar cases seem to produce results that deviate from observable trends.
The fully connected model, although possessing a larger amount of weights by design is surprisingly not amongst the worst performers. It also displays a much more uniform forecasting performance across its hierarchy than its neighboring k-level or cut-tree models.
Both observations can be explained by the fact that it possesses a number of connections in a similar order to k-level and cut-tree partitions while profiting from a more uniform design. This allows the dropout layer to reduce the network in an unconstrained manner, thus functioning under optimal conditions.
Another peculiar behavior can be examined under the tree \textit{td} with \textit{shc} loss which displays few, but impacting, poor performances across its hierarchy, thus negatively affecting its mean accuracy performance.

\subsection{Forecast coherency}
\begin{figure}
    \centering
    \begin{adjustbox}{width=0.70\linewidth}
        \includegraphics{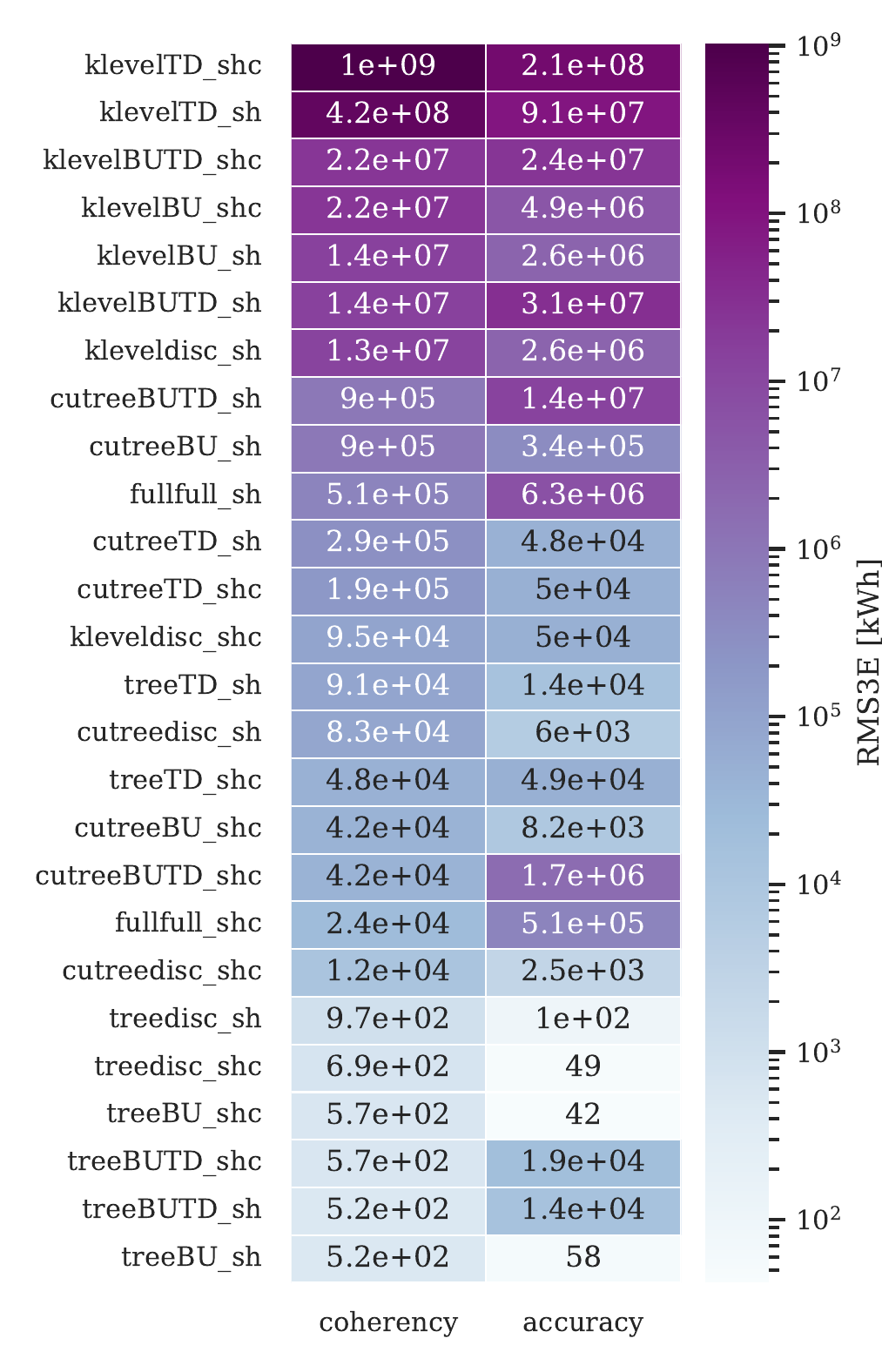}
    \end{adjustbox}
    \caption{Heatmap of the accuracy and coherency Root Mean Structurally-Scaled Square Error (RMS3E) across the forecasting methods. Structural hierarchical forecasting methods described against the y-axis are designated by their respective partition, topological bridge, and considered loss function, and are here sorted according to their forecasted coherency.}
    \label{fig:heatmap_cohacc}
\end{figure}

While information exchange across a hierarchy in a forecasting setting has demonstrated accuracy gain potentials, the coherency improvements of the produced hierarchical time series must be evaluated.
Figure \ref{fig:heatmap_cohacc} subsequently presents the coherency RMS3E, as defined by Eq. \eqref{eq:cohrency_loss}, sorted across evaluated hierarchical model designs. The coherency errors can be compared to their associated accuracy biases, thus providing a complete overview of a method's performance.

The models producing the most coherent forecast range between 5.2e2 and 9.7e2 kWh RMS3E and all benefit from tree partitions, either connected in a \textit{bu}, \textit{butd}, or \textit{disc} fashion, by decreasing order of performance respectively.
These top coherency performers also relate to top accuracy ones, with the exception of the tree-\textit{butd} model.
While both tree-\textit{bu} and tree-\textit{butd} produce slightly more coherent forecasts without the inclusion of coherency information in their loss functions, i.e., \textit{sh}, compared to their coherent counterpart, \textit{shc}, the coherency MS3Es are fairly similar, and theses differences can here be neglected.

The most incoherent forecasts are here produced by models with k-level partitions and \textit{td}, \textit{butd}, or \textit{bu} linkages, ranging between 1e19 and 1.3e7 kWh RMS3Es respectively.
Models including coherency information in their learning process here also display poorer coherency performances but are, however, associated with extremely poor accuracy performances.

A surprising observation showcases the fully connected model, \textit{full}, and cut-tree-\textit{butd} with coherency losses as some of the better coherency performers, in spite of their poor accuracies and large number of weights to learn.
Generally, however, coherency performances display similar tendencies as their associated accuracy ones.


\subsection{Coherency information value}\label{subsect:coherencyval}
\begin{figure}
    \centering
    \subfloat[]{\includegraphics[width = 0.80\linewidth]{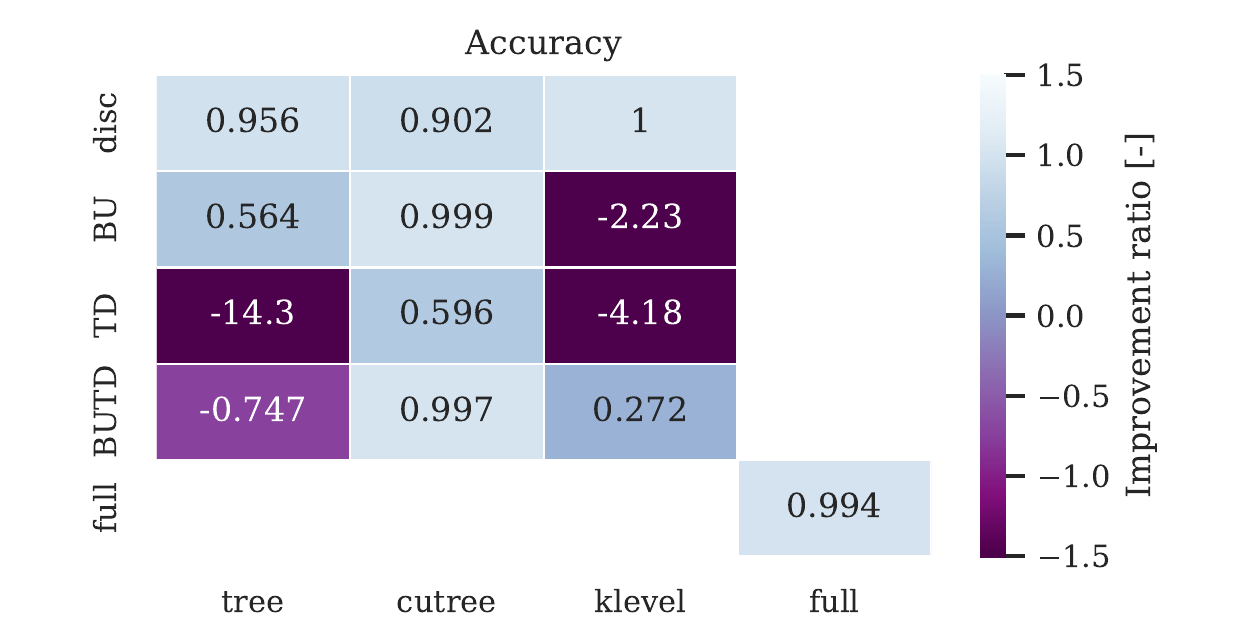}} \\
    \subfloat[]{\includegraphics[width = 0.80\linewidth]{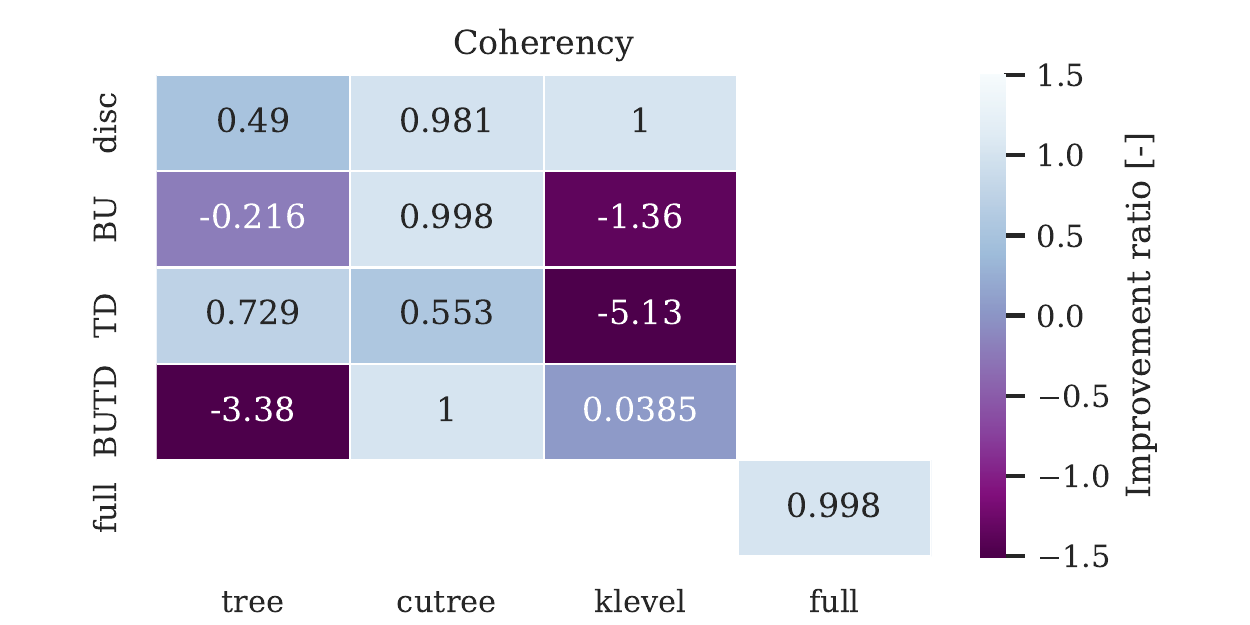}}
    \caption{Value ratio brought by the coherency loss function for (a) the accuracy, and (b) coherency performances of the forecast.}
    \label{fig:accuracy_value}
\end{figure}

Finally, to formally investigate the value brought by coherency information in the learning process of structural-hierarchical models, we evaluate the relative performance ratio between \textit{sh} and \textit{shc} loss functions of similar models.
The improvement ratios for accuracy, \textit{acc}, and coherency, \textit{coh}, are defined as
\begin{flalign}
    r_{acc} = \frac{\mathcal{L}^{sh}-\mathcal{L}^{sh}_{coh}}{\mathcal{L}^{sh}} \text{ ,} \\
    r_{coh} = \frac{\mathcal{L}^{sc}-\mathcal{L}^{sc}_{coh}}{\mathcal{L}^{sc}} \text{ ,}
\end{flalign}
where $r$ is the improvement ratio defined by the difference between structural-hierarchical losses $\mathcal{L}^{sh}$ or structural coherent ones $\mathcal{L}^{sc}$ and their respective counterparts with the inclusion of coherency in the learning mechanism of the model, i.e., Eq. \eqref{eq:Lshc}.
This difference is then normalized by the reference loss, which does not consider coherency information in its loss function, i.e., Eq. \eqref{eq:Lsh}.
As such, positive improvement ratios relate to a performance improvement brought by coherency knowledge, whereas negative ratios point to performance regressions.
This echoes the relative root mean square error (RRMSE) \cite{ATHANASOPOULOS201760} evaluation metric typically employed to estimate the value brought by a reconciliation approach to a base forecast. The main difference in this setting is that instead of a common base forecast, we consider the structural-hierarchical forecast performance from each individual model architecture. This allows a relative performance evaluation per model architecture of the inclusion of coherency information in the learning process of the regressors.

Figure \ref{fig:accuracy_value} presents the improvement ratios categorized by their network design characteristics, i.e., per partition and topological bridge arrangement.
Both accuracy and coherence improvements brought by the coherency loss only display four cases of performance regression, three of which are similar: k-level-\textit{bu}, k-level-\textit{td}, and tree-\textit{butd}.

Models k-level \textit{bu} and \textit{td} are extreme poor performers both in accuracy and coherency and can thus be disregarded in the remainder of the examination, together with k-level \textit{butd}.

The tree-\textit{butd} and tree-\textit{bu} designs are the two best coherency performers, with equivalent 5.2e2 to 5.7e2 kWh RMS3Es.
The accuracy of the tree-\textit{butd} model, however, is more modest, with 1.4e4 and 1.9e4 kWh RMS3E for \textit{sh} and \textit{shc} corresponding losses.
Due to the equivalent, top-performing coherencies of these models, their obtained negative coherency improvement ratios can thus be considered null and consequently disregarded.
The remaining network characteristics all exhibit improved coherency forecasts thanks to the inclusion of the coherency loss in their learning procedure. 
This significant finding places structural hierarchical coherent learning as a valuable method, bringing forecasts one step closer to coherency, prior to reconciliation.

Regarding the accuracy improvements brought by the coherency loss, the tree-\textit{td} design demonstrates an interesting behavior where coherency is improved but accuracy deteriorates.
By looking further into the accuracy performance of this method in Fig. \ref{fig:heatmap_acc}, it was noted that the model produced overall good accuracies across its nodes with the exception of a few extreme cases, which significantly impact the overall performance of the forecast. As such, the \textit{shc} loss function consequently pushes the forecast to a more coherent outcome than its \textit{sh} equivalent, at the cost of a poorer accuracy across the hierarchy.
A similar, but less pronounced, outcome can be observed for the tree-\textit{butd} network, which maintains a similar coherency score but tapers its accuracy by adjusting fewer excessive forecasts.
The coherency value investigation consequently allows us to claim that coherency knowledge improves the accuracy of produced hierarchical forecasts provided individual forecasts are generated within reasonable accuracy limits.

    \section{Conclusion}\label{sect_con}

To secure coherent forecasts across hierarchical structures, recent research proposed hierarchical learning as a pioneer solution bridging the two formerly distinct phases that are forecasting and reconciliation.
While the work depicted promising potential, results displayed disparate performances where coherency information was only found to improve forecasting performances in one setting. 
Additionally, the method suffered from two prevailing challenges, namely, an arduous learning process and faulty coherent learning as a result of input hierarchical time-series normalization.

This work addresses both complications by investigating custom neural network designs echoing the structural topologies of hierarchies.
The approach notably exploits layer partitions producing distinct model components tailored to node-specific elements, while sharing specific information across the model from varying topological bridges resulting in 13 different model architectures. 
Batch normalization is notably included between layers of the model, providing structural-scale robustness to the learning process, while exempting input hierarchical time series from prior normalization. 
We investigate all designs under two novel structurally-scaled learning functions, i.e., structural-hierarchical loss and structural hierarchical-coherent loss, leveraging the mean structurally-scaled square error (MS3E) \cite{https://doi.org/10.48550/arxiv.2301.12967}, and subsequently entitle our approach \textit{structural hierarchical learning}.

The varying neural network designs are evaluated over the accuracy and coherency performances of their produced forecasts from real-world measurements extracted from the Building Data Genome project 2 (BDG2) \cite{miller2020building}.
Models with tree partitionings notably performed best, particularly coupled to bottom-up and disconnected topological bridges, for both structural-hierarchical and structural hierarchical-coherent losses.
Links between the performance of a model and its network topology specifically revealed that (\textit{i}) structural models with fewer connections performed overall better than models with larger numbers of connections, and (\textit{ii}) that the inclusion of coherency information in the loss function improved both the accuracy and coherency performances of forecasts, provided individual forecasts were generated within reasonable accuracy limits.

This work consequently confirms the value potential brought by coherency information in structural hierarchical regressors and places structural hierarchical learning as a successful hierarchical-forecasting method, bringing forecasts one step closer to coherency, prior to reconciliation.
By putting forward tailored, ingenious architectures of neural networks we effectively reduced hierarchical model complexities while serving advanced and coherency-aware hierarchical forecasts.
The approach could notably support domains such as retail, stock management, and distribution networks, thanks to improved and more consistent predictions across all levels of considered hierarchies.

Finally, to encourage knowledge dissemination we render our work fully replicable by open-sourcing all developed python implementations under the public GitHub repository: \href{https://github.com/JulienLeprince/structuralhierarchicallearning}{https://github.com/JulienLeprince/structuralhierarchicallearning}. 


\section{CRediT authorship contribution statement}\label{sect_credit}

\textbf{Julien Leprince}: Conceptualization, Methodology, Software, Data curation, Formal analysis, Visualization, Writing - original draft, review and editing.
\textbf{Waqas Khan}: Conceptualization, Methodology, Writing - review and editing.
\textbf{Henrik Madsen}: Methodology, Supervision, Validation, Writing - review and editing.
\textbf{Jan Kloppenborg M{\o}ller}: Methodology, Supervision, Validation, Writing - review and editing.
\textbf{Wim Zeiler}: Supervision, Funding acquisition.

All authors have read and agreed to the published version of the manuscript.

\section{Acknowledgments}
This work is funded by the Dutch Research Council (NWO), in the context of the call for Energy System Integration \& Big Data (ESI-bida). Additionally, funding related to SEM4Cities (Innovation Fund Denmark, No. 0143-0004) is gratefully acknowledged.

\bibliography{manuscript/src/mybibfile}

\end{document}